\documentclass[]{bytedance}
\usepackage[toc,page,header]{appendix}

\usepackage{minitoc}
\usepackage{amsfonts}
\usepackage{amssymb}
\usepackage{tabularx}
\usepackage{listings}
\usepackage{xcolor}
\usepackage{cancel}

\usepackage{tabulary,multirow,xspace}
\usepackage{fixmath,mathtools,nicefrac,mmstyle}
\usepackage{subcaption}
\usepackage{caption}
\usepackage{wrapfig} % for wrapfigure
\usepackage[misc]{ifsym} % to use \Letter
\usepackage{colortbl}

\usepackage{wrapfig}
\usepackage{multicol}
\usepackage[most]{tcolorbox}
\usepackage{pifont}

\definecolor{codegreen}{rgb}{0,0.6,0}
\definecolor{codegray}{rgb}{0.5,0.5,0.5}
\definecolor{codepurple}{rgb}{0.58,0,0.82}
\definecolor{backcolour}{rgb}{0.95,0.95,0.92}
\definecolor{boxblue}{RGB}{57,89,163}
\definecolor{boxbluebg}{RGB}{230,237,250} 

\lstdefinestyle{mystyle}{
    backgroundcolor=\color{backcolour},   
    commentstyle=\color{codegreen},
    keywordstyle=\color{magenta},
    numberstyle=\tiny\color{codegray},
    stringstyle=\color{codepurple},
    basicstyle=\ttfamily\footnotesize,
    breakatwhitespace=false,         
    breaklines=true,                 
    captionpos=b,                    
    keepspaces=true,                 
    numbers=none,                    
    numbersep=5pt,                  
    showspaces=false,                
    showstringspaces=false,
    showtabs=false,                  
    tabsize=2
}
\definecolor{mygray1}{gray}{.95}
\definecolor{mygray2}{gray}{.9}
\definecolor{mygray3}{gray}{.95}
\usepackage{pifont}% http://ctan.org/pkg/pifont

\newlength\savewidth
\newcolumntype{x}[1]{>{\centering\arraybackslash}p{#1pt}}

\newcommand{\app}{\raise.17ex\hbox{$\scriptstyle\sim$}}

\usepackage{xcolor}
\usepackage{graphicx}
\usepackage{amssymb}
\usepackage{pifont}
\usepackage{floatrow}
\usepackage{amsmath} 
\usepackage{float}
\usepackage{wrapfig}
\usepackage{multirow}
\usepackage{tcolorbox}
\tcbuselibrary{breakable, skins, raster}
\usepackage{listings}
\usepackage{listings}

\definecolor{commentgreen}{rgb}{0.1, 0.4, 0.1}
\definecolor{keywordblue}{rgb}{0.1, 0.1, 0.7}
\definecolor{stringred}{rgb}{0.7, 0.1, 0.1}

\lstdefinestyle{mystyle}{
    commentstyle=\color{commentgreen},
    keywordstyle=\color{keywordblue},   
    stringstyle=\color{stringred},
    basicstyle=\ttfamily\scriptsize, 
    breaklines=true,
    keepspaces=true,
    showstringspaces=false,
    frame=none,                     
    language=Python, 
}

\newcommand{\name}{HuMo}
\title{\name{}: Human-Centric Video Generation via Collaborative Multi-Modal Conditioning}

\author[1,\star]{Liyang Chen}
\author[2,\star]{Tianxiang Ma}
\author[2]{Jiawei Liu}
\author[2,\dagger]{Bingchuan Li}
\author[2]{\\Zhuowei Chen}
\author[2]{Lijie Liu}
\author[1]{Xu He}
\author[2]{Gen Li}
\author[2]{Qian He}
\author[1,\S]{Zhiyong Wu}

\affiliation[1]{Tsinghua University}
\affiliation[2]{Intelligent Creation Lab, ByteDance}
\contribution[\star]{Equal contribution}
\contribution[\dagger]{Project lead}
\contribution[\S]{Corresponding author}

\abstract{
Human-Centric Video Generation (HCVG) methods seek to synthesize human videos from multimodal inputs, including text, image, and audio. Existing methods struggle to effectively coordinate these heterogeneous modalities due to two challenges: the scarcity of training data with paired triplet conditions and the difficulty of collaborating the sub-tasks of subject preservation and audio-visual sync with multimodal inputs.
In this work, we present \textbf{HuMo}, a unified HCVG framework for collaborative multimodal control. 
For the first challenge, we construct a high-quality dataset with diverse and paired text, reference images, and audio.
For the second challenge, we propose a two-stage progressive multimodal training paradigm with task-specific strategies. For the subject preservation task, to maintain the prompt following and visual generation abilities of the foundation model, we adopt the minimal-invasive image injection strategy. For the audio-visual sync task, besides the commonly adopted audio cross-attention layer, we propose a focus-by-predicting strategy that implicitly guides the model to associate audio with facial regions. For joint learning of controllabilities across multimodal inputs, building on previously acquired capabilities, we progressively incorporate the audio-visual sync task. During inference, for flexible and fine-grained multimodal control, we design a time-adaptive Classifier-Free Guidance strategy that dynamically adjusts guidance weights across denoising steps.
Extensive experimental results demonstrate that HuMo surpasses specialized state-of-the-art methods in sub-tasks, establishing a unified framework for collaborative multimodal-conditioned HCVG.
}

\date{\today}

\checkdata[Project Page (Demo, Codes, Models, Data)]{\url{https://phantom-video.github.io/HuMo}}
\begin{document}
\maketitle

\section{Introduction}
\begin{figure}[t]
\centering
\includegraphics[width=0.98\textwidth]{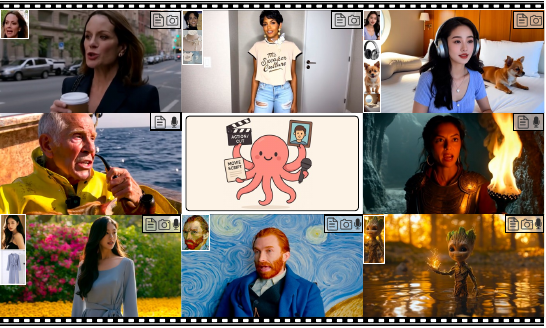}
\caption{We propose \textbf{HuMo}, a multimodal HCVG framework that supports flexible input compositions of text-image (first row), text-audio (second row), and text-image-audio (third row). HuMo generalizes to humans, humans with objects or animals, stylized humanoid artworks, and animations.}
\label{fig:teaser}
\end{figure}
\begin{figure}[t]
\centering
\includegraphics[width=0.98\textwidth]{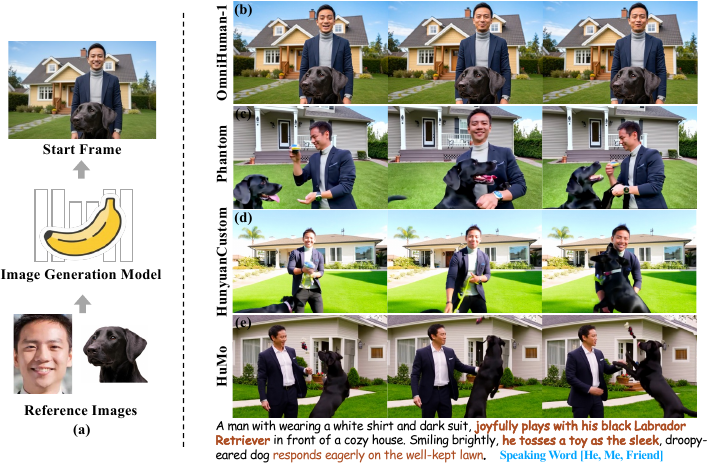}
\caption{Prior HCVG methods comparison. Reference images as inputs for Phantom \cite{phantom_2025}, HunyuanCustom \cite{hunyuancustom_2025}, and HuMo. OmniHuman-1 \cite{omnihuman_2025} takes the start frame as input, which is synthesized by an image generation model \cite{seedream30_2025, banana2025}. OmniHuman-1 suffers from weak text adherence, unable to generate subjects (e.g., a toy) absent in the start frame, while the subject preservation is precariously dependent on the preceding image generator. Phantom lacks audio-driven articulation to synchronize mouth movements with the spoken words in the input audio. HunyuanCustom delivers unbalanced performance on both fronts. HuMo excels in collaborative performance across video quality, subject consistency, audio-visual sync, and text controllability.} 
\label{fig:motivation}
\end{figure}
Recent advances in foundational video generation models have significantly accelerated progress in Human-Centric Video Generation (HCVG), improving both the fidelity and controllability. Such progress is democratizing short video production. Traditional filmmaking, entailing scene setup, casting, styling, and scripting, is labor-intensive and demands collective expertise, incurring substantial time and financial costs. HCVG methods offers a disruptive alternative by leveraging multi-modal inputs: text for describing scenes and actions, images for defining human identity and subject appearance, and audio for character speaking. It greatly reduces manual overhead and enables scalable and efficient content creation.

Prior HCVG methods \cite{hallo3_2025,omnihuman_2025,fantasytalking_2025} typically adopt a two-stage pipeline: a text-to-image (T2I) model \cite{seedream30_2025} generates a subject-complete start frame containing all required elements (e.g., human subject, clothing, accessories or props, and scene background), followed by an image-to-video (I2V)-based model for audio-driven animation. 
However, this pipeline heavily depends on the subject-complete start frame, which inherently constrains the flexibility of text controls for refining or modifying the depicted subjects, as illustrated in Fig.~\ref{fig:motivation}(a)(b).
Alternatively, subject-consistent video generation (S2V) methods \cite{phantom_2025,magref_2025} leverage subject reference images and support flexible video customization through textual prompts. However, these methods are unable to incorporate an audio modality and cannot control what a character is speaking, as shown in Fig.~\ref{fig:motivation}(c). This significantly limits their application in scenarios requiring audio-visual collaboration.
Recently, some methods \cite{hunyuancustom_2025,interacthuman_2025} attempt to integrate the audio-visual sync and subject preservation tasks mentioned above to achieve multimodal control. However, as shown in Fig.~\ref{fig:motivation}(d), they struggle to achieve effective collaboration among the triplet input modalities of text, reference images, and audio. For instance, emphasizing the influence of reference images usually degrades audio-visual sync. Conversely, prioritizing high sync tends to compromise either the text following or the subject preservation ability.

This work focuses on the HCVG task conditioned on text, reference images, and audio, aiming to achieve collaborative multimodal control. There exist two major challenges:
1) Data scarcity. Training such a model requires paired data with precisely aligned triplet conditions. However, there is a lack of publicly available datasets or standardized data preprocessing pipelines to effectively integrate these heterogeneous modalities.
2) Difficulty in collaborative multimodal control. Within a multi-task training framework, it is difficult to simultaneously achieve strong text following, subject consistency with reference images, and accurate audio-visual sync, as these abilities tend to compromise or undermine one another during learning. To tackle these challenges, we propose HuMo. Our key insight lies in the joint design of the data processing pipeline and training paradigm that enables the collaborative learning of multimodal controllability using the constructed multimodal data.

To address the issue of data scarcity, we construct a two-stage multimodal data processing pipeline. Firstly, leveraging large-scale text–video samples \cite{koala_2025,openhumanvid_2025}, we retrieve reference images with the same semantics but different visual attributes \cite{phantom_2025, phantomdata_2025} for each subject in the video samples from a billion-scale image corpus through detection, matching, filtering, and verification. This aims to ensure faithful subject preservation with the reference images while enabling flexible text editability on the subjects. Secondly, to achieve audio-synchronized video generation, we further filter video samples with synchronized audio tracks using speech enhancement and speech-lip alignment estimation \cite{latentsync_2025}. Through this pipeline, we establish a high-quality multimodal dataset containing paired triplet conditions (text, reference images, and audio), which offers a strong foundation for subsequent learning of multimodal controllability.

Building upon the constructed dataset, we propose a progressive multimodal training paradigm consisting of two sub-tasks. 1) The \textbf{Subject Preservation Task} aims to enable collaborative control between text and image by preserving reference images without compromising the text-following capability of the foundational DiT-based backbone \cite{wan_2025}. This is achieved through a minimally invasive image injection strategy that avoids structural modifications to the DiT backbone and confines parameter updates to a limited subset.
2) The \textbf{Audio-Visual Sync Task} introduces audio cross-attention layers to incorporate the audio modality. Unlike previous methods \cite{magicinfinite_2025,fantasytalking_2025} that directly localize audio influence, we propose a focus-by-predicting strategy that implicitly encourages the model to enhance synchronization between audio and human motions. 
To ensure that the previously acquired subject preservation ability will not be compromised during learning of audio-visual sync, we retain the first task while \textbf{progressively incorporating} the second task.
This joint optimization strategy facilitates collaborative learning of controllability across text, image, and audio modalities. By decoupling the learning of individual capabilities, this paradigm also enables flexible training on partially annotated datasets. For instance, enhancing subject preservation only requires video–image pairs, without the need for synced audio–video data.

During inference, to enable flexible and fine-grained control over triplet input modalities, we propose a time-adaptive classifier-free guidance (CFG) strategy, which dynamically adjusts the guidance strengths at different denoising steps. This allows for precise and efficient collaboration among text following, subject preservation, and audio-visual sync.

The main contributions can be summarized as follows:
\begin{itemize}
\item \textbf{Concept.} We attribute the imbalanced multimodal controllability in existing HCVG methods to the absence of a comprehensive design across data processing and training paradigms for handling heterogeneous inputs. We propose HuMo, a unified framework that enables collaborative control across text, image, and audio modalities. It seamlessly supports various input compositions of text-image, text-audio, and text-image-audio.
\item \textbf{Methodology.} 1) We establish a multimodal data processing pipeline that produces a high-quality and diverse dataset with paired triplet conditions.
2) We propose a progressive multimodal training paradigm with task-specific strategies, which facilitates joint learning of controllabilities across triplet modalities. 3) We design a time-adaptive CFG strategy, enabling flexible, fine-grained, and collaborative multimodal control.
\item \textbf{Significance.} Extensive experimental results show HuMo surpasses the specialized state-of-the-art (SOTA) methods on both subject preservation and audio-visual sync sub-tasks. We further demonstrate its effectiveness and scalability by validating the framework on models of 1.7B and 17B parameters.
\end{itemize}
\section{Related Works}
\subsection{Audio-Driven Human Animation}
Audio-driven human animation aims to generate videos from input human images to produce lip movements matching input speech signals.
Built on the foundational video generation models \cite{hunyuanvideo_2024, wan_2025, seedance_2025, seawead_2025, goku_2025}, audio-driven human animation methods have achieved impressive performance.
Hallo3 \cite{hallo3_2025} is the first full-frame-size portrait animation application on a pre-trained DiT model, using cross-attention layers for audio control. To improve the motion dynamics, FantasyTalking \cite{fantasytalking_2025} proposes to establish global motion at the clip level and optimize lip synchronization at the frame level.
OmniHuman-1 \cite{omnihuman_2025} scales the training data by incorporating hybrid motion-related conditions to generate more realistic human videos. 
Despite their strong performance in facial animation and body motion, existing methods still require a subject-complete start frame with visible facial features, which limits user creativity.

\subsection{Subject-Consistent Video Generation}
Subject-consistent video generation (S2V) aims to generate text-aligned videos with consistent subject appearances according to the input text prompts and reference images. Early approaches use pre-trained semantic encoders \cite{clip_2021} to extract features from reference images, achieving identity-preserving video generation via adapters \cite{Idanimator_2024, identity_2025} or DiT cross-attention layers \cite{moviegen_2024}.
In-context methods \cite{phantom_2025, magref_2025} leverage the inherent consistency of pre-trained DiT-based models by concatenating reference image latents with noisy video latents.
These methods achieve finer-grained subject consistency but weaken textual control.
To preserve the textual control of the pre-trained model in our in-context method, we freeze the text-visual cross-attention layers and fine-tune only the self-attention layers. 
Moreover, existing methods support only image and text modalities and cannot support human speech input, which is critical for vivid and realistic video creation. 
Concurrent works such as InterActHuman \cite{interacthuman_2025} and HunyuanCustom \cite{hunyuancustom_2025} integrate subject-consistent video generation with audio-driven human animation.
However, they still struggle to balance the influence of these three modalities. In contrast, we propose a progressive multimodal collaborative training paradigm and time-adaptive CFG to enable flexible multimodal control.

\section{Methodology}
\begin{figure*}[t]
\centering
\includegraphics[width=\textwidth]{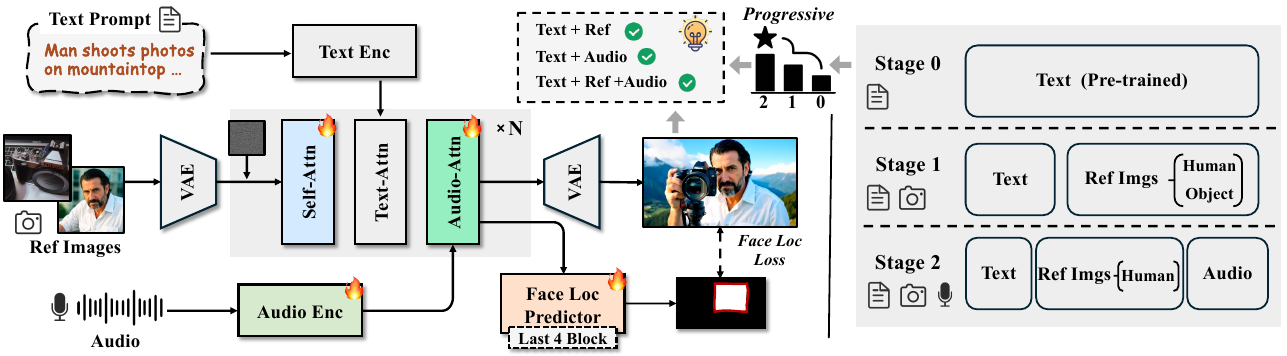}
\caption{\textbf{Overview of our framework.} HuMo model (left) is trained based on the proposed data processing pipeline (right). Built upon a DiT-based T2V backbone from Stage 0, the model progressively learns subject preservation and audio-visual sync capabilities in Stages 1 and 2. HuMo achieves collaborative generation across different modality compositions.}
\label{fig:pipeline}
\end{figure*}
To address the challenges of data scarcity and compromised performance in multimodal conditioning, we propose \textbf{\uline{HuMo}}, a \textbf{\uline{Hu}}man-Centric Video Generation framework that enables collaborative control with \textbf{\uline{M}}ultim\textbf{\uline{o}}dal conditions.
Given a textual prompt $c_\text{txt}$, reference images $c_\text{img}$, and an audio singal $c_\text{a}$, HuMo aims to generate a video where the environment, human identity, appearance, accessories, clothing, as well as facial and body motions are semantically aligned with the input conditions, while remaining spatially and temporally coherent.
We begin in Sec.~\ref{sec:preliminaries} by describing the DiT-based T2V backbone \cite{wan_2025}. Sec.~\ref{sec:data_process} outlines the construction of the multimodal dataset. In Sec.~\ref{sec:collaborative_train}, we present how a T2V model is extended to support triplet modalities. Finally, Sec.~\ref{sec:inference} describes our inference strategy for flexibly modality conditioning and fine-grained and collaborative generation.

\subsection{Preliminaries}
\label{sec:preliminaries}
In this work, we adopt a DiT-based T2V model \cite{wan_2025} as our backbone.
As shown in Fig.~\ref{fig:pipeline}, it incorporates a 3D Variational Autoencoder (VAE) to compress video into a compact latent space. A text encoder \cite{umt5_2023} is utilized to encode textual information and then injected into the DiT backbone with cross-attention. For training, flow matching \cite{flowmatching_2023} is adopted by learning a continuous velocity field that transports samples from a simple prior to the data distribution along deterministic trajectories. HuMo inherits this DiT-based architecture and extends it by incorporating additional image and audio modalities, enabling multimodal conditioning for more controllable video generation. To effectively train the model under this multimodal setup, we formulate the flow matching objective as:
\begin{equation}
\mathcal{L}_{\mathrm{FM}}(\theta)=\mathbb{E}_{t,z_0,z_1}\|v_\theta(z_t,t,c)-(z_1-z_0)\|_2^2,
\label{eq:flow_matching_loss}
\end{equation}
where $z_0$ is a sample from the simple prior,  $z_1$ is the latent of the target video sample, and $c= \{{c}_{\text{txt}}, c_{\text{img}}, c_\text{a}\}$ denotes the multimodal condition of text, image, and audio. The DiT model $v_\theta$ takes the noisy latent $z_t = (1 - t){z}_0 + tz_1$, and learns to predict its velocity at any point $t\in[0,1]$. This formulation allows HuMo to efficiently learn a deterministic transport map from noise to data under complex multimodal constraints.

\subsection{Multimodal Data Processing Pipeline}
\label{sec:data_process}
High-quality, subject-consistent, and audio-visual synced multimodal data is crucial for HCVG but remains scarce and usually incomplete or misaligned. We propose a multimodal data processing pipeline to construct a diverse dataset of triplet conditions, which forms the foundation for the training paradigm we introduce later.

As shown in Fig. \ref{fig:pipeline}, our pipeline unfolds in several stages. In Stage 0, we begin with a large-scale video pool \cite{koala_2025,openhumanvid_2025} and leverage powerful VLMs \cite{qwen25vl_2025, gemini25_2025} for detailed text descriptions, ensuring basic textual modality for each video sample. In Stage 1, to avoid the common copy-paste issue \cite{phantom_2025} in subject preservation, we follow prior works \cite{phantomdata_2025,opens2v_2025} to sparsely sample video frames and retrieve cross-paired reference images from a billion-scale image corpus. For humans, we retrieve images with the same identity but varying in appearance (e.g., makeup, pose, background, clothing, age); for objects, we match the same semantic category but with varied visual attributes (e.g., color, shapes, viewpoint). This strategy improves text controllability for reference-guided generation by discouraging direct frame replication in training.
In Stage 2, we supplement the data with audio modality by filtering for speech segments and creating tightly aligned audio-visual pairs with lip-sync analysis \cite{latentsync_2025}. It is worth noting that in Stage 2, since audio-aligned data predominantly exists in human-centric videos, we can only retrieve human reference images. 

Through this pipeline, we build a multimodal dataset consisting of detailed text prompts, consistent reference images, synced audio, and the target video. Stage 1 includes $\mathcal{O}(1)$M video samples with text and reference images, while Stage 2 contains $\mathcal{O}(50)$K samples with temporally-aligned audio. We will release the training data to facilitate future research on high-quality, controllable HCVG and to promote reproducibility in multimodal generation studies.

\subsection{Progressive Multimodal Training}
\label{sec:collaborative_train}
We structure the acquisition of multimodal control capabilities into two distinct yet progressive training stages. Stage 1 establishes text-image controllability through the subject preservation task. Stage 2 realizes the joint learning of text-image-audio controllability via progressive training of the previous task and the audio-visual sync task. Training data for each stage is sourced from the corresponding stage of our data processing pipeline.

\noindent \textbf{Subject Preservation.}
In Stage 1, we introduce reference images as additional inputs. To preserve the strong text following and image synthesis ability of the original DiT backbone \cite{wan_2025}, we adopt the minimal-invasive image injection strategy, which adheres to two key principles: avoiding structural modifications of the DiT backbone and confining parameter updating to a limited subset.

Specifically, without modifying the model architecture, we concatenate the VAE latents $z_{\text{img}}$ of the reference images $c_{\text{img}}$ with the noisy latent $z_t$ along the temporal dimension as inputs. $c_{\text{img}}$ can be a single image of a human or an object, or the composition of multiple images.
To prevent the model from misinterpreting the reference as the start frame and performing undesired image continuation, we always place the reference latents at the end of the video latent sequence, forming the input as $[z_t; z_{\text{img}}]$.
This design encourages the model to actively extract subject identity information from the reference images via self-attention and propagate it across all frames, enabling subject-consistent generation. The training is restricted to the self-attention layers of DiT to minimally affect its inherent text alignment and visual generation capabilities. Notably, text remains as input to ensure semantic consistency and controllability in generation.

\noindent \textbf{Audio-Visual Sync.}
In stage 2, we extend the first-stage setup by incorporating audio modality. Following prior works \cite{fantasytalking_2025, multitalk_2025}, we insert an audio cross-attention layer in each DiT block.
Audio features are extracted via Whisper \cite{whisper_2023} for generalization across speakers and languages.
Based on the observation that human motions are primarily aligned with temporally nearby audio cues \cite{syn_obama_2017},  we construct the final audio embedding $c_\text{a} \in \mathbb{R}^{f\times n \times d}$ by concatenating audio embeddings within a temporal window centered at each video frame stamp, where the window length $n=5$ and and $f$ same with the sequence length of the video latents.
$c_\text{a}$ is conducted cross-attention calculation with hidden states frame-by-frame:
\begin{equation}
    \operatorname{Attention}(h_z, c_\text{a})=\operatorname{softmax}(\frac{\mathbf{Q}_z \mathbf{K}_\text{a}^{\top}}{\sqrt{d}}) \mathbf{V}_\text{a},
\end{equation}
where $\mathbf{Q}_z$ transformed from the hidden states $h_z \in \mathbb{R}^{f\times h \times w\times d}$ of video latent in DiT net, $\mathbf{K}_a$ and $\mathbf{V}_a$ transformed from audio embedding.

The aforementioned cross-attention is computed between the audio signal and the full-frame noisy latent. As is known, audio cues are usually most correlated with localized human regions (e.g., face, lip). Consequently, this coarse-grained attention can lead to weak synchronization between audio and human motion. Prior I2V-based methods \cite{magicinfinite_2025} address this by detecting facial regions in the start frame and restricting audio cross-attention exclusively to these areas. This strategy, however, is not applicable when the input is a full-frame facial image, as the model cannot predetermine the location of the face before denoising. 

We therefore propose a focus-by-predicting strategy to implicitly guide the model to focus more on the facial region. During training, we introduce a mask predictor $\mathcal{F}_\text{mask}$ to estimates the potential facial region distribution $\mathbf{M}_\text{pred}=\mathcal{F}_\text{mask}(h_z)$. $\mathbf{M}_\text{pred}$ is supervised by the ground-truth binary face mask $\mathbf{M}_\text{gt}$ using a binary cross-entropy (BCE) loss.
Since early DiT blocks lack stable spatial representations, we only insert $\mathcal{F}_\text{mask}$ after the audio cross-attention modules in the last four DiT blocks and average these outputs as the final predicted mask. 
To further prevent the supervision from weakening when the face region is too small, we introduce a size-aware weight \cite{magicinfinite_2025} by adaptively emphasizing the supervision on the face region:
\begin{equation}
    \mathcal{L}_\text{mask} = \frac{hw}{\sum_{i=1}^{h} \sum_{j=1}^{w} \mathbf{M}_\text{gt}^{(i,j)}} \cdot \text{BCE}(\mathbf{M}_\text{pred}, \mathbf{M}_\text{gt}).
\end{equation}

Unlike prior methods \cite{magicinfinite_2025,interacthuman_2025} that apply hard gating on audio attention outputs, which is a truncation design we consider suboptimal. The focus-by-predicting strategy acts as a soft regularizer, which steers the model's focus without crippling its representational capacity and retains the flexibility to model full-body kinematics and complex interactions.

\begin{figure}[t]
\centering
\includegraphics[width=0.85\textwidth]{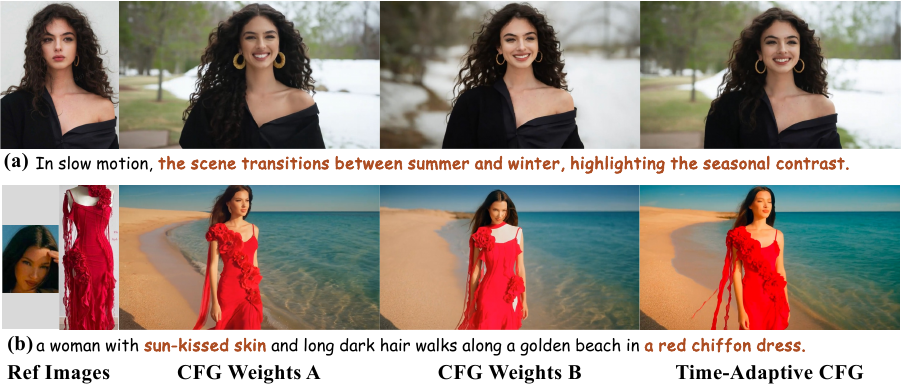}
\caption{The proposed \textbf{time-adaptive CFG} balances text guidance and identity preservation.}
\label{fig:cfg}
\end{figure}
\noindent \textbf{Progressive Training.}
To ensure that the acquisition of audio-visual sync ability does not degrade the subject preservation ability, we employ a progressive task-weighting curriculum in stage 2. Initially, training is dominated by the subject preservation task (80\% ratio, audio input as null) to reinforce existing ability, while the audio-visual sync task constitutes the remaining 20\%. As training progresses, we gradually increase the proportion of the audio-visual sync task to 50\%.
Throughout this stage, we still adhere to the finetuning principles in stage 1 by only updating the self-attention layers and audio-related modules.
This progressive strategy facilitates a smooth transition for the model from bi-modal (text, image) to tri-modal (text, image, audio) inputs, ensuring stable training and the collaboration of multimodal learning. 

\subsection{Inference Strategies}
\noindent \textbf{Flexible Multimodal Control.} For flexible, independent control at inference, we adapt CFG with separate guidance scales ($\lambda_\text{txt}$, $\lambda_\text{img}$, $\lambda_\text{a}$) for each modality:
\begin{align}
v_\theta(z_t,t,c) & = \lambda_\text{a}[v_\theta(c_\text{txt}, c_\text{img}, c_\text{a})-v_\theta(c_\text{txt}, c_\text{img}, \varnothing)] \notag\\
& + \lambda_\text{img}[v_\theta(c_\text{txt}, c_\text{img}, \varnothing)-v_\theta(c_\text{txt}, \varnothing, \varnothing)] \notag\\
& + \lambda_\text{txt}[v_\theta(c_\text{txt}, \varnothing, \varnothing)-v_\theta(\varnothing, \varnothing, \varnothing)] \notag\\
& + v_\theta( \varnothing, \varnothing, \varnothing).
\end{align}
HuMo can generate coherent results even when modalities are absent, enabling condition combination of  $[c_\text{txt}, c_\text{img}]$, $[c_\text{txt}, c_\text{a}]$, and $[c_\text{txt}, c_\text{img}, c_\text{a}]$. Missing conditions are simply replaced by a null token $\varnothing$.

\noindent \textbf{Time-Adaptive CFG.}
\label{sec:inference}
We observe that the influence of each modality shifts throughout the denoising process: 
early steps tend to construct the overall semantic structure and spatial layout guided by text, while later steps focus on fine-grained details (e.g., identity similarity, audio-visual sync). A static CFG configuration is suboptimal for the shifting modal dynamics. 
We therefore propose a time-adaptive CFG that dynamically switches between two configurations of guidance scales. From timestep 1.0 to 0.98, we adopt the text/image-dominant configuration to establish a stable semantic layout (e.g., character and scene composition). From timestep 0.98 to 0, we shift to parameters that emphasize audio and image control. Empirical results demonstrate that this strategy significantly improves multimodal collaboration and enhances overall video quality.
\section{Experiment}
\begin{figure*}[t]
\centering
\includegraphics[width=\textwidth]{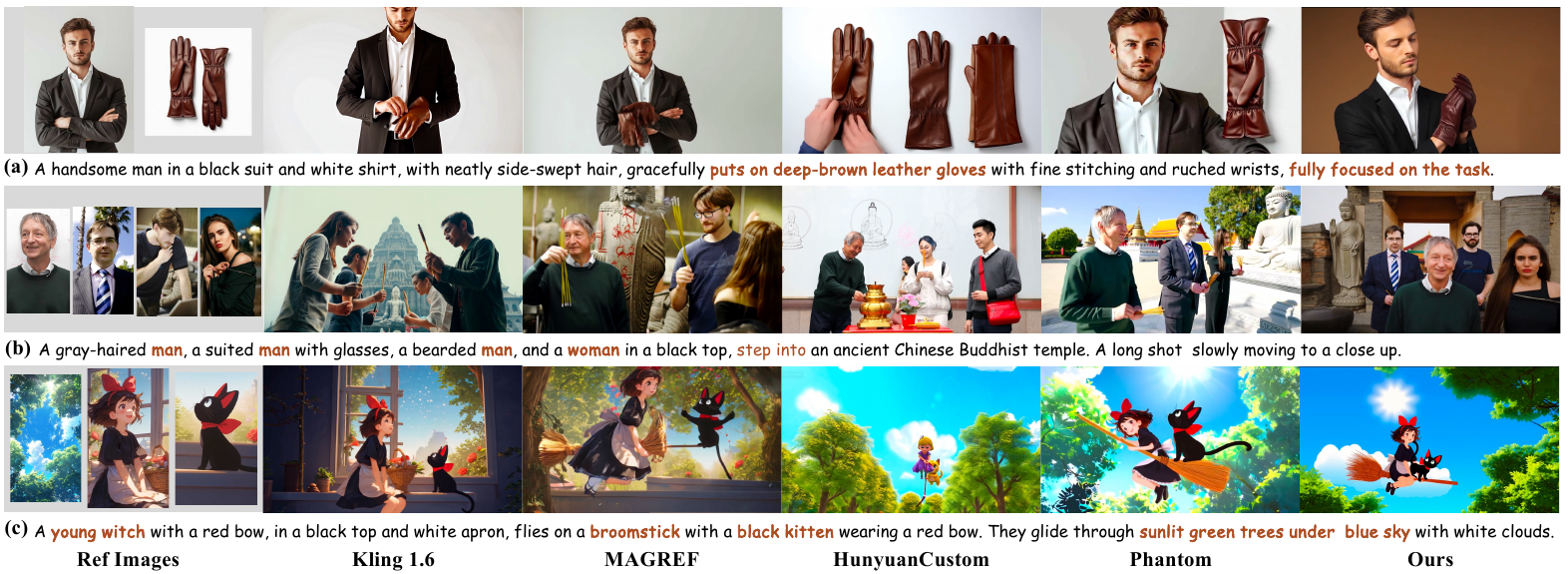}
\caption{\textbf{Qualitative comparison for the subject preservation task.} Zoom in for details.}
\label{fig:s2v_result}
\end{figure*}
\begin{table*}[t]
\centering
\resizebox{\textwidth}{!}{
\small
\begin{tabular}{lcccccccc}
\toprule
\multirow{2}{*}{Method} & \multicolumn{3}{c}{Video Quality}                   & Text Following  & \multicolumn{4}{c}{Subject Consistency} \\
\cmidrule(lr){2-4} \cmidrule(lr){5-5} \cmidrule(lr){6-9}
                        & AES$\uparrow$          & IQA$\uparrow$            & HSP$\uparrow$             & TVA$\uparrow$             & ID-Cur$\uparrow$          & ID-Glink$\uparrow$        & CLIP-I$\uparrow$          & DINO-I$\uparrow$         \\ \midrule
Kling 1.6 \cite{Kling_2025}               & 0.645          & 0.714          & 3.792           & 2.564          & 0.470            & 0.501          & 0.639           & 0.394          \\
MAGREF \cite{magref_2025}                 & 0.622          & 0.708          & 3.331          & 2.852           & 0.334          & 0.359          & 0.665          & 0.416          \\
HunyuanCustom \cite{hunyuancustom_2025}          & 0.592          & 0.705          & 3.705          & 1.777          & 0.309         & 0.335         & 0.649          & 0.426          \\
Phantom \cite{phantom_2025}                & 0.608          & 0.150          & 3.612         & 2.877          & 0.649          & 0.674          & 0.677          & 0.426          \\ \midrule
Ours-1.7B               & 0.586          & 0.680          & 3.432          & 3.222          & 0.609          & 0.668          & 0.660          & 0.414          \\
Ours-17B                & \textbf{0.657} & \textbf{0.717} & \textbf{3.906} & \textbf{3.939} & \textbf{0.731} & \textbf{0.757} & \textbf{0.687} & \textbf{0.447} \\ \bottomrule
\end{tabular}}
% \vspace{-4pt}
\caption{\textbf{Quantitative results for the subject preservation task} with text and reference images as inputs. }
\label{tab:s2v_result}
\end{table*}

\begin{figure*}[t]
\centering
\includegraphics[width=\textwidth]{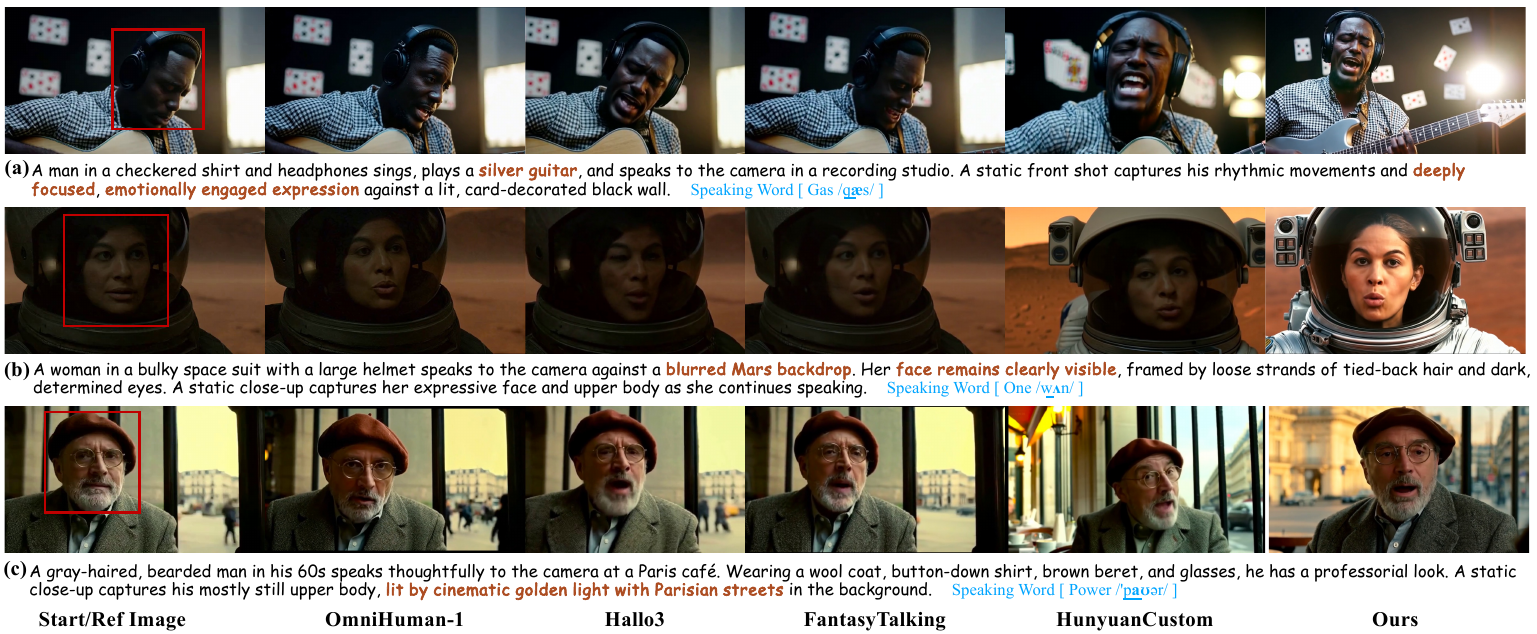}
\caption{\textbf{Qualitative comparison for audio-visual sync task.} For the I2V-based methods, we use the first frame generated by MoCha \cite{mocha_2025} as the input start frame. For HunyuanCustom and Ours, the cropped face from the start frame is utilized as input. The official OmniHuman-1 \cite{omnihuman_2025} website API does not support text input. Zoom in for details.}
\label{fig:avsync_result}
\end{figure*}
\begin{table*}[t]
\centering
\resizebox{\textwidth}{!}{
\small
\begin{tabular}{lcccccccc}
\toprule
\multirow{2}{*}{Method} & \multicolumn{3}{c}{Video Quality}                   & Text Following  & \multicolumn{2}{c}{SubjectSubject Consistency} & \multicolumn{2}{c}{Audio-Visual Sync} \\ \cmidrule(lr){2-4} \cmidrule(lr){5-5} \cmidrule(lr){6-7} \cmidrule(lr){8-9}
                        & AES$\uparrow$     & IQA$\uparrow$ & HSP$\uparrow$ & TVA$\uparrow$             & ID-Cur$\uparrow$                 & ID-Glink$\uparrow$             & Sync-C$\uparrow$       & Sync-D$\downarrow$         \\ \midrule
OmniHuman \cite{omnihuman_2025}               & 0.545          & 0.682            & 4.503           & -               & 0.677                  & 0.727                 & \textbf{6.526}    & \textbf{7.784}    \\
Hallo3 \cite{hallo3_2025}                  & 0.381          & 0.634            & 4.200           & 6.117           & 0.726                  & 0.727                 & 5.189             & 9.212             \\
FantasyTalking \cite{fantasytalking_2025}         & 0.455          & 0.652            & 4.444           & 6.209           & 0.703                  & 0.729                 & 3.202             & 10.914            \\
HunyuanCustom \cite{hunyuancustom_2025}          & 0.358          & 0.619            & 4.370           & 6.246           & 0.729                  & 0.716                 & 4.562             & 9.892             \\ \midrule
Ours-1.7B               & 0.322          & 0.661            & 4.350           & 5.865          & 0.721                  & 0.729                 & 6.005             & 8.648             \\
Ours-17B                & \textbf{0.589} & \textbf{0.718}   & \textbf{4.537}  & \textbf{6.508} & \textbf{0.747}         & \textbf{0.740}        & 6.252             & 8.577             \\ \bottomrule
\end{tabular}}
\caption{\textbf{Quantitative results for the audio-visual sync task} on MoCha benmark.}
\label{tab:avsync_result}
\end{table*}
\noindent \textbf{Implementation Details.} 
We build our method on the backbones of Wan-2.1-1.3B and Wan-2.1-14B \cite{wan_2025}. All video samples are resampled to 25 fps with a 480$\times$832 resolution. We employ a two-stage training strategy. Stage 1 involves training for 40k steps, with all audio-related modules disabled. Stage 2 continues for another 40k steps, where we enable the audio-related modules.

\noindent \textbf{Comparative Methods.} 
We compare HuMo with two categories of baselines. \textbf{1) S2V methods} that take text and subject reference images as input, including open-source models MAGREF \cite{magref_2025}, HunyuanCustom \cite{hunyuancustom_2025}, Phantom-Wan-14B \cite{phantom_2025}, and the commercial closed-source method Kling 1.6 \cite{Kling_2025}. Comparison is conducted on an in-house benchmark of 100 test cases involving humans, objects, and animals.
\textbf{2) Audio-visual synced methods} that take text, image, and audio as inputs, including open-source methods Hallo3 \cite{hallo3_2025}, FantasyTalking \cite{fantasytalking_2025}, HunyuanCustom \cite{hunyuancustom_2025}, and the commercial closed-source OmniHuman-1 \cite{omnihuman_2025}. Comparison is conducted on the MoCha benchmark \cite{mocha_2025}. Notably, HunyuanCustom is the only method that supports reference images, while all other baselines rely on a subject-complete start frame and follow the I2V generation paradigm. The latter methods are easier to achieve higher visual quality.

\noindent \textbf{Evaluation Metrics.}
We conduct comprehensive evaluation using multiple objective metrics covering four key aspects.
\textbf{1) Video Quality.} We evaluate visual appeal and perceptual quality using aesthetics (AES) and image quality assessment (IQA) from the widely-used VBench \cite{vbench_2024}. Specifically for human videos, we leverage the SOTA VLM, Gemini-2.5-Pro \cite{gemini25_2025}, to estimate the human structure plausibility (HSP).
\textbf{2) Text-Video Alignment (TVA).} Semantic consistency between the input text prompt and generated video is measured via the VLM-based reward model \cite{videoalign_2025}.
\textbf{3) Subject Consistency.} For human identity, we detect and crop faces from generated frames, and compute similarity with the reference image using Face-Cur \cite{facecur_2020} and Face-Glink \cite{facecur_glink}. For non-facial objects, we use DINO-I \cite{groundingdino_2024} and CLIP-I \cite{clip_2021} scores to compute average embedding similarity.
\textbf{4) Audio-Visual Sync.} We adopt Sync-C and Sync-D \cite{latentsync_2025} to quantify alignment between input audio and facial motion.

\subsection{Multimodal Conditioned Comparison}
\noindent \textbf{Subject Preservation Task.}
\textbf{1) Qualitatively.} As shown in Fig.~\ref{fig:s2v_result}, HuMo demonstrates superior performance in text following ability compared to other methods. For instance, in case (b), where the prompt is ``step into a temple", other methods fail to generate the correspondence. Our method shows strong subject preservation and generalizes well to unseen four-subject cases. It accurately maintains four distinct human identities in case (b), while other methods suffer from missing persons or human identity drift. 
HuMo also excels in visual aesthetics and human structure plausibility. In case (a), HuMo generates a person wearing gloves without structural artifacts, whereas baselines show noticeable limb degradation. In case (c), HuMo seamlessly integrates a background image into the generation, despite not being trained on any background images.
\textbf{2) Quantitatively.} Tab.~\ref{tab:s2v_result} reveals that HuMo outperforms other methods in aesthetic and overall image fidelity. HuMo achieves a strong capability in modeling human pose and body integrity with the highest HSP score. For text following and subject consistency with reference images, HuMo also achieves the SOTA performance, which is consistent with the qualitative observation. 
This highlights our model's capabilities to achieve strong textual editability without compromising subject consistency.

\noindent \textbf{Audio-Visual Sync Task.} 
\textbf{1) Qualitatively.} Fig.~\ref{fig:avsync_result} first reveals HuMo's superior capability in text following. In case (a) and (c), HuMo successfully synthesizes the ``silver guitar" and ``golden light in the background" respectively, whereas other methods fail to render these specific details. 
This observation underscores a fundamental weakness of I2V-based methods - their limited ability for re-editing the provided subject-complete start frame. Furthermore, in case (b), when provided with a dimly lit facial image, other methods are unable to generate the ``clearly visible face" as required by the text prompt. HuMo, by contrast, not only synthesizes a clear face but also preserves the identity of the reference face image.
\textbf{2) Quantitatively.} As shown in Tab.~\ref{tab:avsync_result}, HuMo attains the highese scores for aesthetic quality and text follwing. In terms of HSP and identity similarity, our method outperforms HunyuanCustom. Notably, HuMo also surpasses other I2V-based methods, despite their inherent advantage with stronger priors on body layout and facial structure.
For audio-visual sync, our 1.7B model already outperforms several open-source specialized models and trails only slightly behind the commercial method OmniHuman-1. It is crucial to note that for this evaluation, we only utilize a single reference face image. As illustrated in Fig.~\ref{fig:teaser}, HuMo supports a combined input of audio and multiple reference images  (e.g., facial photos, clothing, animal), offering enhanced controllability and customization.

\begin{table}[t]
\centering
% \vspace{-0.5em}
\resizebox{0.6\textwidth}{!}{
\small
\begin{tabular}{lcccc}
\toprule
Variants        & AES $\uparrow$            & TVA $\uparrow$           & ID-Cur $\uparrow$        & Sync-C $\uparrow$        \\ \midrule
Full Finetune   & 0.529          & 6.157          & \textbf{0.749} & 6.250          \\
w/o Progressive Training & 0.541          & 6.375          & 0.724          & 6.106          \\
w/o Focus-by-Predicting    & 0.587          & 6.507          & 0.730          & 5.946          \\ \midrule
Ours-17B        & \textbf{0.589} & \textbf{6.508} & 0.747          & \textbf{6.252} \\ \bottomrule
\end{tabular}}
% \vspace{-4pt}
\caption{\textbf{Quantitative ablation study} on MoCha.}
\label{tab:ablation}
\end{table}
\begin{figure}[t]
\centering
\includegraphics[width=0.85\textwidth]{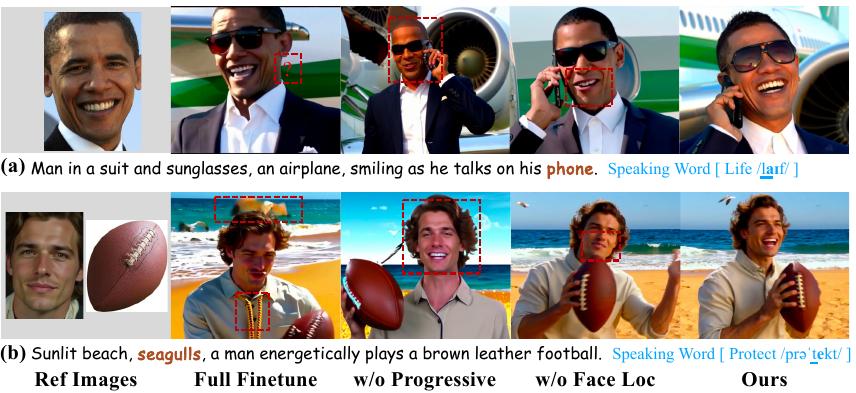}
\caption{\textbf{Qualitative ablation study.} Zoom in for details.}
\label{fig:ablation}
\end{figure}
\subsection{Method Analysis}

\textbf{Ablation Study for Training Strategies.} 1) Full Finetuning. We update all parameters of the DiT model instead of confining parameter updating to a limited subset. It leads to significant drops of AES and TVA scores in Tab.~\ref{tab:ablation}. Visually, Fig.~\ref{fig:ablation}(a) fails to generate the ``phone", and case (b) exhibits noticeable artifacts. Full finetuning disrupts the DiT's pretrained capabilities for high-quality video synthesis and text-image alignment.
2) w/o Progressive Training. We train two tasks in a single stage. This change leads to a degradation across most evaluation metrics. The generated character exhibits low identity similarity to the reference image. This suggests that without progressively building different capabilities, the model struggles to achieve effective coordination across modalities.
3) w/o Focus-by-Predicting. We remove the focus-by-predicting strategy. It results in a decline in Sync-C score, and the generated lip movements are misaligned with the spoken word. 
This highlights the importance of face location prediction for learning audio-visual correspondence. Furthermore, we observe a decline in ID-Cur score, indicating that this strategy also implicitly contributes to better facial identity consistency.

\textbf{Ablation Study for Inference Strategy.} As shown in Fig.~\ref{fig:cfg}, CFG parameters A focus on text-driven layout control, while CFG parameters B emphasize identity similarity. Our proposed time-adaptive CFG dynamically adjusts the weights of the two parameters at different generation stages: in the early time steps, it prioritizes textual guidance to ensure the generated content aligns with the prompt structure; in later time steps, it increases the emphasis on identity preservation. This adaptive strategy effectively balances semantic controllability and identity consistency, leading to more precise and stable character generation.

\begin{figure}[t]
\centering
\includegraphics[width=0.98\textwidth]{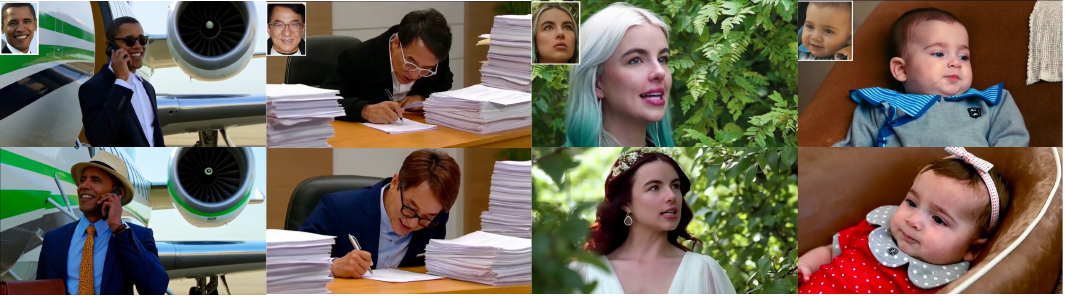}
\caption{Given the same reference subject reference image but different text prompts, our method achieves \textbf{collaborative text-image controllability}.}
\label{fig:text_control}
\end{figure}

\begin{figure}[t]
\centering
\includegraphics[width=0.98\textwidth]{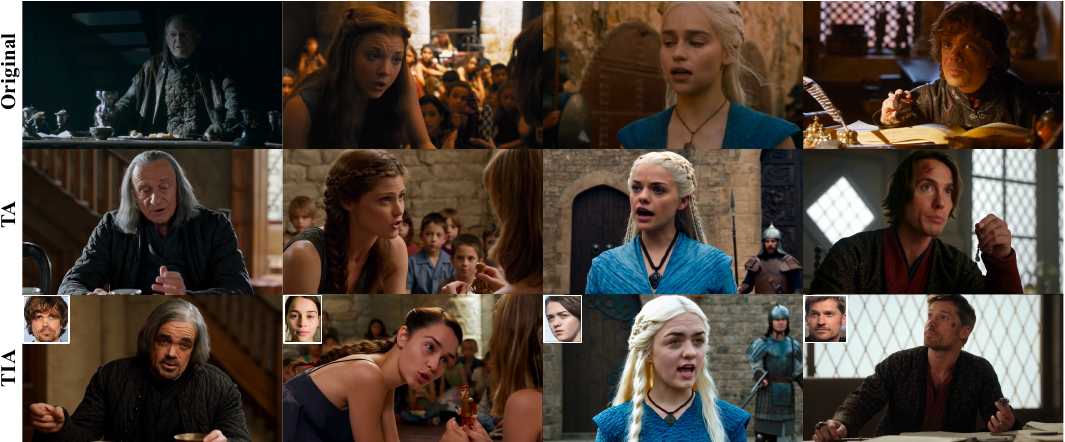}
\caption{\textbf{Re-creation of Game of Thrones (Season 3), named as Faceless Thrones.} We extract captions \cite{qwen25vl_2025} and audio from the original video and generate new videos with HuMo using two modes: \textbf{text–audio (TA)} and \textbf{text–image–audio (TIA)}. The reference identity image for TIA mode is displayed at the top-left corner.}
\label{fig:img_control}
\end{figure}
\textbf{Text controllability.}
As shown in Fig.~\ref{fig:text_control}, we input the same reference image and use different text prompts to vary the character’s clothing, accessories, and makeup in order to evaluate text controllability. We observe that while the character’s identity remains consistent, their appearance changes accordingly, demonstrating the effectiveness of our method in achieving text-image collaboration. Notably, previous subject-preserving methods \cite{phantom_2025, hunyuancustom_2025} primarily focus on compositionality, which embeds reference images within the semantic context of the text, while overlooking the ``editability" of the text, i.e., the ability to adjust the degree of information injected from the reference image. In this work, we specifically enhance this aspect.

\textbf{Image controllability.}
As shown in Fig.~\ref{fig:img_control}, to showcase the movie-level, highly customizable generation capabilities of HuMo, we re-create scenes from the renowned TV series “Game of Thrones” using both text-audio (TA) and text-image-audio (TIA) modes. In the TIA setting, to demonstrate flexible controllability of reference images, we condition the model on a reference portrait of an actor different from the one in the original footage. Across both TA and TIA modes, HuMo produces videos that preserve the similar layout and visual elements of the original scenes. In TIA mode, the reference identity is seamlessly integrated into the target semantic context, enabling short-form production from a single, unadorned headshot of the actor.
\section{Conclusion}
We propose HuMo, a novel human-centric video generation framework with multimodal conditions. HuMo establishes a multimodal data processing pipeline to produce high-quality and diverse datasets with paired text prompts, reference images, and audio. The proposed progressive multimodal training paradigm successfully integrates the control capabilities of text, image, and audio modalities into a unified model. Taking advantage of the proposed time-adaptive CFG strategy, our model enables flexible, fine-grained, and collaborative control over all three modalities during inference. HuMo satisfies the multiple requirements of text prompt following, subject preservation, and audio-visual sync in human-centric short video creation.

\section{Ethical Considerations}
\label{sec:ethical}
The development of HuMo for Human-Centric Video Generation may raise several ethical concerns. First, the ability to synthesize realistic human videos from multimodal inputs (text, image, and audio) may lead to misuse, such as deepfakes or non-consensual content creation. Ensuring informed consent and protecting individuals' likenesses are critical. Second, fine-grained control over generated content calls for responsible usage guidelines to prevent manipulation or misinformation. Developers and users must adhere to ethical standards, including transparency, data privacy, and the prevention of harm.

\clearpage

\bibliographystyle{plainnat}
\bibliography{main}
\end{document}